\documentclass[letterpaper, 10 pt, conference]{ieeeconf}  
\UseRawInputEncoding

\IEEEoverridecommandlockouts                              
\usepackage{amsmath} 
\usepackage{amsfonts} 
\usepackage{amssymb}  
\usepackage{graphicx}
\usepackage{hyperref}
\usepackage{mathtools}
\usepackage{comment}
\usepackage{algorithmicx}
\usepackage{algpseudocode}
\usepackage[table]{xcolor}

\usepackage{xcolor}
\usepackage{caption}
\usepackage{subcaption}
\usepackage[linesnumbered, ruled,vlined]{algorithm2e}

\usepackage[font=footnotesize,labelfont=footnotesize]{subcaption}
\usepackage[font=footnotesize,labelfont=footnotesize]{caption}

\SetKwInput{KwInput}{Input}                
\SetKwInput{KwOutput}{Output}              

\usepackage{amssymb}
\usepackage{amsmath, bm}
\usepackage{verbatim}

\usepackage{multirow}
\usepackage{hhline}
\renewcommand{\vec}{\mathbf}

\usepackage{cite}

\usepackage{color}

\usepackage[normalem]{ulem}

\usepackage{tabularx}

\newcommand{\hysconfig}{\boldsymbol{\eta}}
\graphicspath{{figures/}}

\definecolor{D}{RGB}{196, 114, 0}
\definecolor{Q}{RGB}{18, 113, 148}

\title{\LARGE \bf
Accounting for Hysteresis in the Forward Kinematics of Nonlinearly-Routed Tendon-Driven Continuum Robots via a Learned Deep Decoder Network}

\author{Brian Y. Cho$^{1}$, Daniel S. Esser$^{2}$, Jordan Thompson$^{1}$, Bao Thach$^{1}$, \\Robert J. Webster III$^{2}$, and Alan Kuntz$^{1}$
\thanks{$^{1}$Brian Y. Cho, Jordan Thompson, Bao Thach, and Alan Kuntz are with the Robotics Center and the Kahlert School of Computing at the University of Utah, Salt Lake City, UT 84112, USA; (email: \{brian.cho, alan.kuntz\}@utah.edu).}
\thanks{$^{2}$Daniel S. Esser and Robert J. Webster III are with the Department of Mechanical Engineering, Vanderbilt University, Nashville, TN, 37203.}
\thanks{This material is based upon work supported in part by the National Science Foundation under grant number 2133027. D. Esser was also partially supported by the Natural Sciences and Engineering Research Council of Canada (NSERC) under grant 521537544. Any opinions, findings, and conclusions or recommendations expressed in this material are those of the authors and do not necessarily reflect the views of the NSF or NSERC. (\textit{Corresponding author: Brian Y. Cho.})}
}

\usepackage{color}

\begin{document}

\maketitle
\begin{abstract}
Tendon-driven continuum robots have been gaining popularity in medical applications due to their ability to curve around complex anatomical structures, potentially reducing the invasiveness of surgery. 
However, accurate modeling is required to plan and control the movements of these flexible robots.
Physics-based models have limitations due to unmodeled effects, leading to mismatches between model prediction and actual robot shape.
Recently proposed learning-based methods have been shown to overcome some of these limitations but do not account for hysteresis, a significant source of error for these robots.
To overcome these challenges, we propose a novel deep decoder neural network that predicts the complete shape of tendon-driven robots using point clouds as the shape representation, conditioned on prior configurations to account for hysteresis.
We evaluate our method on a physical tendon-driven robot and show that our network model accurately predicts the robot's shape, significantly outperforming a state-of-the-art physics-based model and a learning-based model that does not account for hysteresis.
\end{abstract}

\vspace{0.2cm}
\begin{keywords}
Continuum robots, Tendon-driven continuum robots, Machine learning-based modeling, Learned kinematics
\end{keywords}

\section{Introduction}
\label{section:intro}

Continuum robots have garnered significant attention for medical applications due to a high degree of dexterity and their flexible nature---enabling them to perform precise and intricate movements in complex and narrow anatomical structures~\cite{Dupont2022_ProcIEEE, Burgner2015_TRO}. 
The inherent flexibility of continuum robots has the potential to minimize the invasiveness of surgery, mitigating one of the problems of open surgery--the overall trauma to the patient. 
Tendon-driven continuum robots are a type of continuum robot actuated by tendons routed along the length of the backbone~\cite{Nguyen2015_IROS, Kato2015_TMECH}. 
When the tendons are pulled, they apply forces and moments to the backbone, causing the robot to bend and take curved shapes, based on its backbone compliance and the tendon routing.

\begin{figure}[t]
    \captionsetup{skip=-0.2pt}
    \centering
    \includegraphics[width=\linewidth]{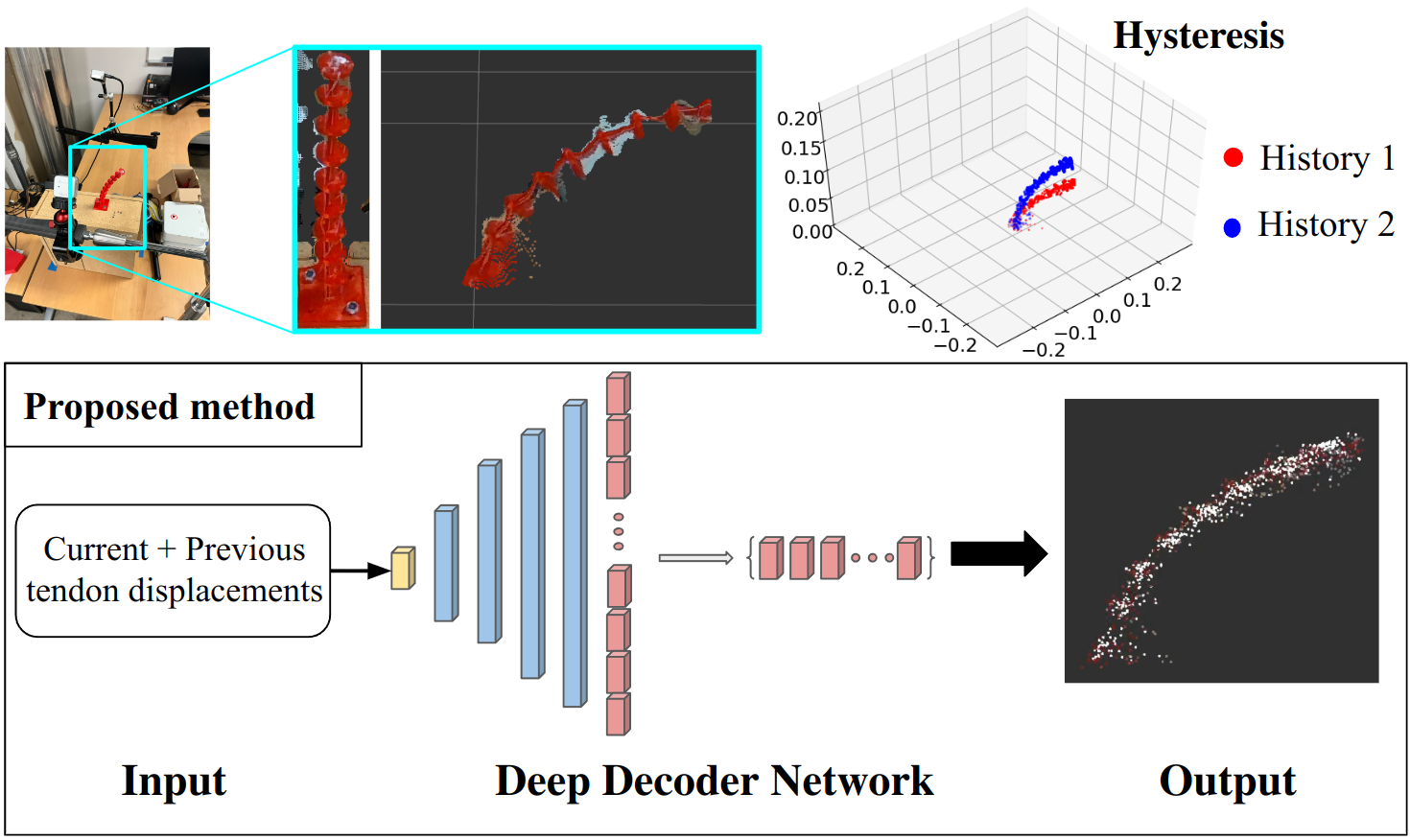}
  \caption{(Upper left) We leverage point clouds to represent the whole shape of the robot. (Upper right) Hysteresis causes the robot's shape to significantly depend on its prior configuration history. History $1$ and History $2$ show the sensed shape of the robot at the same tendon-displacement configuration, but having come from different prior configurations. (Bottom) Our proposed deep decoder neural network model aims to learn the tendon robot's entire shape while accounting for hysteresis. Taking as input the current and previous tendon configurations it produces a point cloud (white) that well aligns with the ground truth (red).}
  \label{fig:intro}
\end{figure}

Accurate mechanical models are necessary to plan and control the complex movements of these flexible robots. 
Physics-based mechanical modeling approaches for tendon-driven robots, e.g.,~\cite{Rucker2011_TRO, Oliver-Butler2019ContinuumDisplacements}, are unable to take into account unmodeled effects such as anisotropic or non-homogeneous material properties and unpredictable friction, leading to a mismatch between model predictions and the actual shape of a physical robot.
Machine learning-based models, e.g.,~\cite{Grassmann2018_IROS,Kuntz2020_TMRB}, have been shown to be able to learn from data to predict robot shape well.
However, to date these methods only map a single current configuration to the robot's shape.
Continuum robots, including tendon-driven robots, suffer from hysteresis, in which the shape is dependent not only on the current configuration but on prior configurations~\cite{Yu2018_MS, Kim2021_arXiv} (see Fig.~\ref{fig:intro}).
These concerns make it difficult to predict a robot's shape for either physics-based or learned methods, which in turn can lead to inaccurate planning and control of these robots.

In this work, we propose a deep neural network-based method that learns to predict the entire shape of a tendon-driven robot given both its current configuration, i.e., tendon displacements, as well as the prior configuration.
Conditioning the shape model on both the robot's current configuration and its previous configuration enables the model to account for hysteresis and significantly improves the accuracy of the shape prediction.

As an additional benefit, our method outputs a geometrical representation of the robot's shape \emph{directly} in the form of a point cloud---a representation widely used in other domains for object geometric shape representation~\cite{Wu2019_CVPR, Qi2017_CVPR,Thach2022_ICRA}.
This is in contrast to existing methods, e.g.,~\cite{Kuntz2020_TMRB}, which output parameterizations of shape such as the robot's backbone.
These parameterizations require additional, time-consuming computation to describe the robot's full shape in order to be useful for control or planning via collision detection or task-space metric evaluation. 
Our method's point-cloud output is enabled by our deep-decoder-based neural network architecture as well as a novel point-cloud-based loss function.

We evaluate the effectiveness of our proposed method in improving the shape prediction accuracy on a physical tendon-driven robot with both linear and non-linearly routed tendons. 
We demonstrate significant improvement in shape prediction compared to a state-of-the-art physics-based mechanical model.
We also demonstrate the method's ability to successfully account for hysteresis by comparing it against a version that is not conditioned on prior configurations. 
The results show that our deep decoder network model accurately predicts the robot's shape and successfully accounts for hysteresis.

\section{Background and Related Work}

Tendon-driven robots are a class of continuum robots that use cables or tendons to control their motion.
They have been widely used in various applications, including surgical robotics~\cite{Burgner2015_TRO, Dupont2022_ProcIEEE, Bentley2022_IA, Kato2015_TMECH}. 
Tendon-driven robots display high dexterity and flexibility, enabling a large workspace and the ability to navigate through confined spaces and curve around anatomical obstacles in the human body.
Additionally, they can be designed to be lightweight and compact, which makes them suitable for applications where size and weight are a concern~\cite{Swaney2016_JMD, Leavitt2023_Hamlyn}. 
One of the major challenges in tendon-driven robots is accurately modeling their complex mechanical behavior.
This is largely due to non-linear and difficult to model factors such as hysteresis, friction, and anisotropic material properties, resulting in inaccuracies in predicting the shape and motion of tendon-driven robots.

To address this challenge, research effort has been made to handle many unmodeled effects in tendon-driven robots~\cite{Yuan2019_MMT, Yu2018_MS, Kim2021_arXiv, Poignonec2020_RAL}.
Of particular concern is a phenomenon displayed by tendon-driven robots called hysteresis~\cite{Yu2018_MS, Kato2016_IJCARS, Kim2021_arXiv, baek2020hysteresis, guo2023motion}. 
This manifests as a dependence in the forward kinematics computation not only on current configuration but on past configurations or paths.
Hysteresis occurs in tendon-driven robots due to multiple potential factors, e.g., friction between the tendons and the robot, actuator slack, and tension irregularities.

Researchers have proposed various modeling techniques for continuum robots, from physics-based models, both analytical and numerical~\cite{Yuan2019_MMT, Rucker2011_TRO, Neppalli2009_AR}, to machine learning-based approaches~\cite{Kuntz2020_TMRB, Grassmann2018_IROS, wu2021hysteresis, bai2021task}.
Rucker \textit{et al.}~\cite{Rucker2011_TRO} present a physics-based model that leverages Cosserat rod theory to model the mechanics of tendon-driven continuum robots. 
Tendon friction is a known issue in such robots that can introduce hysteretic behaviour in these devices, and various approaches have been proposed to model such effects in catheter-like devices with straight tendons \cite{Jung2011_IROS, Subramani2015_ICRA}, however modeling these effects in generally routed (i.e., not necessarily linearly-routed) continuum robots is an open challenge.
Lilge \textit{et al.}~\cite{Lilge2022_IJRR} propose a method for state estimation of concentric tube robots using Gaussian process regression that enables inferring a continuum robot's shape and internal strain variables from noisy sensor data.
Data-driven methods have been leveraged to compute the inverse kinematics~\cite{Xu2017_IJMRCAS, Liang2021_ICRA} and forward kinematics~\cite{Bergeles2015_Hamlyn, Grassmann2018_IROS, Fagogenis2016_IROS} of continuum robots.
However these methods do not consider the full robot shape, rather only computing information about the robot's tip---which may be insufficient for applications requiring, e.g., obstacle avoidance.
Kuntz \textit{et al.}~\cite{Kuntz2020_TMRB} propose a forward kinematics method for learning the entire backbone shape of concentric tube robots, however the method does not directly output the robot shape.
Further, none of these learned methods account for hysteresis.
We build upon these methods in this work.

\section{Problem Formulation}

We consider a tendon-driven robot consisting of a flexible backbone of length $l$ with $N$ tendons arbitrarily, e.g., potentially \textit{non-linearly}~\cite{Rucker2011_TRO}, routed along its length, with reference base frame $\mathcal{B}$.
By pulling a tendon $i$ at the robot base, the compliant backbone bends, and the entire robot changes its shape according to the tendon routing and the tendon displacement defined by $q_i$ for tendon $i$.
We then define the robot's configuration as a vector $\vec{q} = [q_i : i = 1,\cdots,N] \in \mathbb{R}^{N}$.
Furthermore, to enable the consideration of hysteresis, we define an augmented \emph{hysteresis configuration} vector $\hysconfig$ which pairs the prior robot configuration along with the current configuration, $\hysconfig = [\vec{q}_\mathrm{prior}, \vec{q}_\mathrm{current}] \in \mathbb{R}^{2N}$.

We define the robot's shape as $\vec{p}$, noting that $\vec{p}$ can take the form of any sufficiently descriptive shape representation.
We next define a function that maps the robot configuration, conditioned on the prior configuration, to the robot's shape as the forward kinematics function $\mathcal{F}_{FK} : \hysconfig \rightarrow \vec{p}$. 
The problem then becomes to approximate $\mathcal{F}_{FK}$ as closely as possible. 

\section{Method}

To solve this problem, our method leverages a learning-based approach to map hysteresis configurations to the robot's shape in the form of a point cloud.
Specifically, we propose a novel deep decoder neural network model that maps an $\hysconfig = [\vec{q}_\mathrm{prior}, \vec{q}_\mathrm{current}] $ to a point cloud $\hat{\vec{p}}$ consisting of $M$ 3D points such that $\hat{\vec{p}} \in \mathbb{R}^{3 \times M}$,
representing the complete geometry of the robot.
The goal is to output $\hat{\vec{p}}$ such that it closely approximates the true shape of the robot $\vec{p}$ after having started in $\vec{q}_\mathrm{prior}$ and then transitioned to $\vec{q}_\mathrm{current}$, i.e., having executed the two configurations in $\hysconfig$ in sequence.

\begin{figure}[t]
    \captionsetup{skip=-0.2pt}
    \centering
    \includegraphics[width=\linewidth]{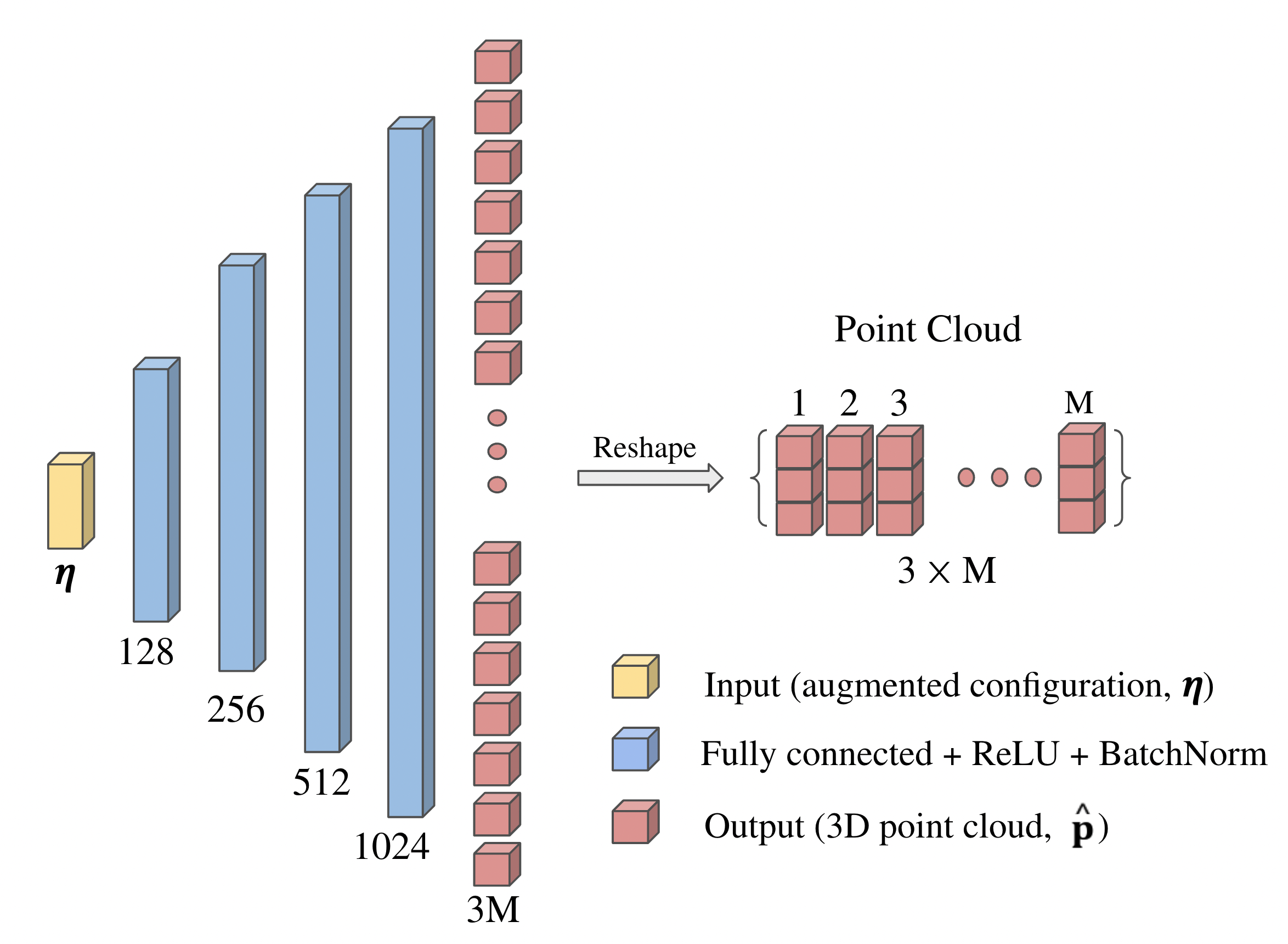}
  \caption{Network architecture of our novel deep decoder network. The model takes the robot's augmented configuration vector $\hysconfig$ as input and outputs a point cloud of the robot's shape $\hat{\vec{p}}$. The model consists of $4$ hidden layers, each of which is fully-connected, ReLU activated, and batch normalized followed by an output layer of size $3M$. We reshape the network output (1-dimensional vector) into a $3 \times M$ matrix, where $M$ is the number of points, as the point cloud representation.}
  \label{fig:network_architecture}
\end{figure}

Our deep decoder network model architecture is presented in Fig~\ref{fig:network_architecture}.
The model takes as input $\hysconfig$.
This is then passed through four fully-connected layers of increasing size, each with rectified linear unit (ReLU) activation functions~\cite{Nair2020_ICML} and batch normalization~\cite{Ioffe2015_ICML}.
This is then connected to an output layer defining a 1-dimensional vector of size $3M$.
We reshape the network output into the point cloud as a $3 \times M$ matrix, a general form of point clouds, where $M$ is the number of points.

We present a novel loss function $\mathcal{L}_{tendon}$ that enables our method to learn the complex and detailed 3D geometry of the tendon-driven robot. $\mathcal{L}_{tendon}$ blends two popular correspondence-free point cloud distance metrics, Chamfer distance and Earth Mover's Distance (EMD). 
Chamfer distance $C$ measures the distance between two sets of points (e.g., $\vec{p}_a$ and $\vec{p}_b$) by summing the distances between each point in one set and its closest neighbor in the other set: 
\[C(\vec{p}_a, \vec{p}_b) = \sum_{x\in \vec{p}_a}\min_{y\in \vec{p}_b} ||x - y||^2 + \sum_{y\in \vec{p}_b}\min_{x\in \vec{p}_a} ||x - y||^2 .\]
By contrast, EMD, defined here as the function $E$, measures the distance between two point distributions via:
\[E(\vec{p}_a, \vec{p}_b) = \min_{\xi: \vec{p}_a \rightarrow \vec{p}_b}\sum_{x\in \vec{p}_a}||x - \xi(x)||_2,\]
where $\xi$ is a bijection pairing points in $\vec{p}_a$ with points in $\vec{p}_b$.
Intuitively, EMD computes the minimum amount of work required to transform one point set into the other, where the work represents the amount of mass moved times the distance it is moved. 
Our novel loss function is the linear combination, with scaling factor $\lambda$ (which we set experimentally as discussed in Sec.~\ref{section: training}), of these two distance metrics between the predicted and ground-truth point clouds: \[ \mathcal{L}_{tendon} = C + \lambda \cdot E.\]  

Intuitively for our problem, Chamfer distance encourages the coarse geometry of our model's predicted point clouds to be similar to that of the ground truth while EMD ensures that the predicted point clouds have a similar distribution to the ground truth.
Adding EMD on top of Chamfer distance helps refine the 3D geometry as well as enforce the points to be evenly distributed on the surface of the tendon robot.

\section{Data collection and model training}
\begin{figure*}[t!]
    \captionsetup{skip=-0.2pt}
    \centering
    \includegraphics[width=\linewidth]{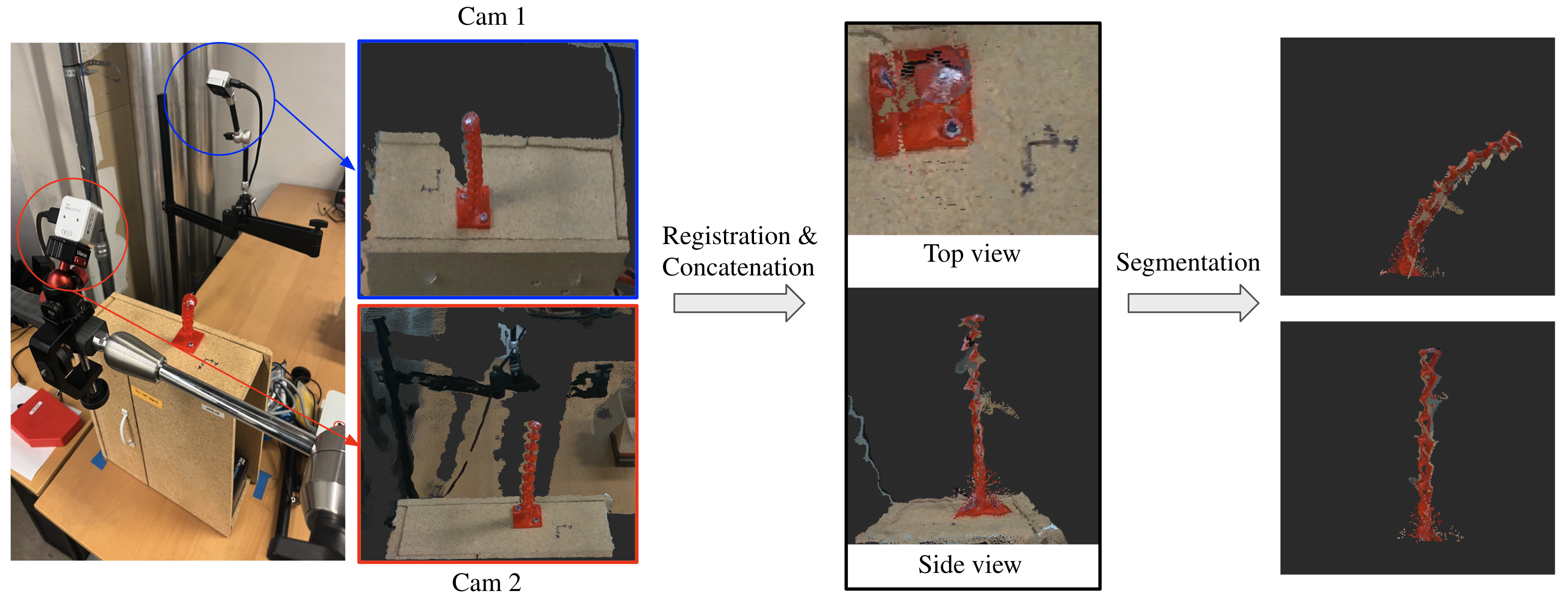}
  \caption{Registration and Segmentation. (Left) We sense point clouds of the robot's shape using two RGB-D cameras placed around the robot. (Middle) We apply a registration algorithm to align the two camera frames with the robot base frame. The concatenated, aligned point clouds are shown. (Right) We then segment the robot's point cloud by removing point cloud points outside of the robot's workspace. 
  }
  \label{fig:data_processing}
\vspace{-1.5em}
\end{figure*}
To successfully train our method, we must first collect a diverse set of hysteresis configurations and pair each with a dense and accurate point cloud representing the shape of the robot.
In this section we first describe how we capture dense and accurate point clouds of the robot's shape and the process by which we segment the robot in the point cloud, separating it from the background.
We next describe the process by which we collect a data set of diverse hysteresis configurations using the point cloud capturing method.
Finally we describe the training process for our model that leverages the data set.

\subsection{Registration and Segmentation}
\label{section:camera_calibration}

To collect point cloud data for a given robot shape, we leverage two point cloud sensors pointed at the robot from different angles (see Fig.~\ref{fig:data_processing}).
These sensors collect the point clouds $\vec{p}_\mathrm{cam1}$ and $\vec{p}_\mathrm{cam2}$\footnote{
In a slight overload of notation we borrow the use of $\vec{p}$ (and associated subscripts) to now also represent a sensed point cloud that lies on the surface of the physical robot.}.
We seek to unify these two point clouds into a single dense point cloud.
To do so, we must register the point clouds to each other as well as to the robot.
Accurate registration of the robot shape from two different camera views requires calibration of the coordinate systems of the cameras to the robot base frame $\mathcal{B}$.
To do so, we leverage the commonly-used correspondence-based registration algorithm of~\cite{Arun1987_TPAMI}.
We identify five points of correspondence in each of the three frames, specifically the robot's tip when it is at its home, zero-tension position, as well as four points at the corners of the 3D printed structure at the robot's base.
The registration algorithm then provides a closed-form solution to generate the optimal rigid transform based on the correspondence set.

We compute two homogeneous transformation matrices $\mathbf{T}_1^{\mathcal{B}} : \vec{p}_\mathrm{cam1} \rightarrow \mathcal{B}$ and $\mathbf{T}_2^1 : \vec{p}_\mathrm{cam2} \rightarrow \vec{p}_\mathrm{cam1}$.
We then calculate the transformation matrix $\mathbf{T}_2^{\mathcal{B}} = \mathbf{T}_1^{\mathcal{B}} \mathbf{T}_2^1$.
Using the resulting transformation matrices, we align the two point clouds $\vec{p}_\mathrm{cam1}$ and $\vec{p}_\mathrm{cam2}$ with the robot base frame $\mathcal{B}$ (see Fig~\ref{fig:data_processing}).

After calibrating the coordinate systems of the two cameras to the robot base frame, we segment the tendon-driven robot, removing the points associated with the background and leaving only the points in $\vec{p}_\mathrm{cam1}^{\mathcal{B}}$ and $\vec{p}_\mathrm{cam2}^{\mathcal{B}}$ that correspond to the robot. 
Specifically, we eliminate any points in the sensed point clouds which are outside of the robot's workspace, which we approximate for this purpose as a hemisphere of radius $l$ (corresponding to the robot's backbone length) plus a safety margin of $\epsilon$ centered at the robot's base. 
This ensures that only the points corresponding to the robot's shape are retained in the scene. 
With the two camera frames registered and the robot segmented, we can concatenate $\vec{p}_\mathrm{cam1}^{\mathcal{B}}$ and $\vec{p}_\mathrm{cam2}^{\mathcal{B}}$ to obtain a complete representation of the robot's shape from both camera views, i.e., $\vec{p} = [\vec{p}_\mathrm{cam1}^{\mathcal{B}}, \vec{p}_\mathrm{cam2}^{\mathcal{B}}]$ (see Fig~\ref{fig:data_processing}).
Note that while we use two cameras, the process above could be used with just one, or with more than two with minimal changes.

\subsection{Hysteresis configuration data set generation}
\label{section:training_details}
To train our model in a supervised fashion where the predicted outputs are compared against the ground-truth point clouds, we collect a data set $\mathcal{D}^{hys}$.
$\mathcal{D}^{hys}$ is a set of data points where each data point is a tuple pairing a hysteresis configuration $\hysconfig$ with its corresponding registered, segmented, and concatenated point cloud $\vec{p}$ collected on the robot.
More formally, $\mathcal{D}^{hys} = \{(\hysconfig_1, \vec{p}_1), (\hysconfig_2, \vec{p}_2), \cdots, (\hysconfig_L, \vec{p}_L)\}$ is a data set of $L$ data points.

Remembering that each $\hysconfig$ encodes two robot configurations (a $\vec{q}_\mathrm{prior}$ and a $\vec{q}_\mathrm{current}$), we note that for the model to learn how hysteresis influences the robot's shape at a given $\vec{q}_\mathrm{current}$, we need $\mathcal{D}^{hys}$ to contain multiple hysteresis configurations where $\vec{q}_\mathrm{current}$ is consistent but $\vec{q}_\mathrm{prior}$ varies.

The process by which we generate this diverse data set is specific to our robot.
The robot on which we evaluate our method (see the red-colored robot in Figs.~\ref{fig:intro} and~\ref{fig:data_processing}) is actuated via four tendons, three straight tendons and a helical tendon (see full description in Sec.~\ref{sec:exp-details}).

We first generate a nominal set of configurations for our training set, generated via a discretization over the combinations of possible tendon displacements on the robot.
We restrict this set to combinations that ensure that all three straight tendons are never tensioned simultaneously (which would not be done in practice) and that the helical tendon is not tensioned simultaneously with any straight tendon.

We utilize this nominal set as the set of possible $\vec{q}_\mathrm{current}$ configurations.
We then generate a large set of $\hysconfig$ augmented configurations in the following way.
First, consider a home configuration as a special case $\vec{q}_{\mathrm{home}}$, where all tendons have zero tension.
We create the first set of hysteresis configurations as $\hysconfig = [\vec{q}_{\mathrm{home}}, \vec{q}_\mathrm{current}]$ for each $\vec{q}_\mathrm{current}$ configuration in our nominal set.
In other words, for each configuration in the nominal set, we first command the robot to the home configuration, $\vec{q}_{\mathrm{home}}$, and then to the non-home configuration $\vec{q}_\mathrm{current}$.
We then record the point cloud at $\vec{q}_\mathrm{current}$.

We next augment this set with more hysteresis configurations where the robot does \emph{not} return to $\vec{q}_{\mathrm{home}}$ between configurations but rather first visits a different, randomly selected configuration from the nominal configuration set.
In other words, for each nominal configuration, we first command the robot to go to a randomly selected configuration and then to the nominal configuration, recording the point cloud at the nominal configuration.
This ensures that we have diverse hysteresis configurations, i.e., that for each configuration in the nominal set (i.e., each $\vec{q}_\mathrm{current}$), we have multiple corresponding hysteresis configurations in which $\vec{q}_\mathrm{prior}$ is different but $\vec{q}_\mathrm{current}$ is the same.

\subsection{Model training}
\label{section: training}

To train and validate our model, we split the full data set into $50$ test cases and separate $70\%$ of the remaining cases for training and $30\%$ for validation. 
After each training epoch, the model is evaluated on the validation set.
If the validation loss stops improving over $50$ consecutive training epochs, we stop the training process to prevent overfitting.
We initialize the model weights using Xavier initialization~\cite{Glorot2010_AISTATS} and use a batch size of $32$ during training. 
We use the Adam optimizer with a learning rate of $0.01$, decayed by a factor of $0.1$ every $100$ epochs.
We use our loss function $\mathcal{L}_{tendon}$ described prior.
By training and evaluating multiple versions of our model, we have determined that $\lambda = 1$  works well.

\section{Experiments and Results}
In this section we present experimental details, quantify how significant hysteresis is for our system, analyze the benefit of our novel loss function, and finally compare the performance of our model to a state-of-the-art physics-based model as well as to a version of our model that does not account for hysteresis. Also note that, with the exception of model training, throughout this section when we compare two point clouds---e.g., a model-predicted point cloud and a ground truth point cloud---we employ Chamfer distance as a metric of difference.
We utilize Chamfer distance rather than the custom loss function we train the model with to ensure the evaluation is not compounding the model performance and loss function performance.

\subsection{Experimental details}
\label{sec:exp-details}

\begin{figure}[t]
    \captionsetup{skip=-1pt}
    \centering
    \includegraphics[width=0.95\linewidth]{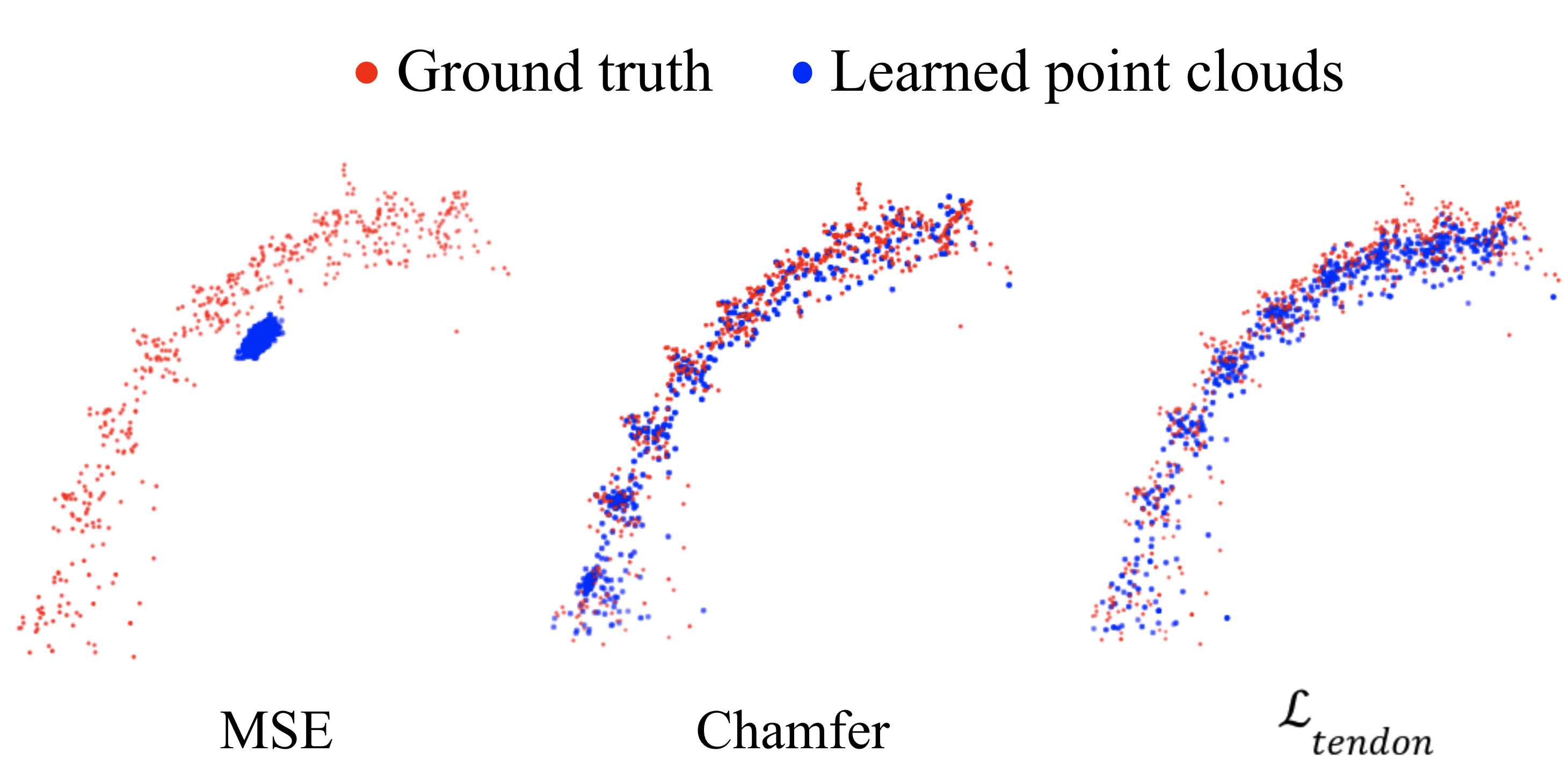}
  \caption{Qualitative loss function comparison. 
  The sensed, ground truth robot shape point cloud is shown in red, and the model-predicted point clouds are shown in blue. 
  The model trained with MSE loss completely fails to achieve the robot's shape. 
  The model trained with the Chamfer loss function produces point clouds that show the robot's geometry but the resulting points are unevenly distributed along the shape. 
  Our proposed loss function, EMD + Chamfer ($\mathcal{L}_{tendon}$), shows the best performance both qualitatively and quantitatively, demonstrating a precise estimation of the robot's shape with evenly distributed points.}
  \label{fig:loss_functions}
 \vspace{-1.5pt}
\end{figure}

For our experiments we leverage a physical tendon-driven robot, composed of a 3D printed, flexible TPU material body with a thin nitinol tube embedded in the 3D printed structure with a length of $0.2$\,m, consisting of $9$ circular disks that connect $3$ straight-routed tendons at $120$ degrees apart and $1$ helically-routed tendon, with linear actuators pulling on the tendons to control the robot's shape at the robot's base (see Figs.~\ref{fig:intro} and~\ref{fig:data_processing}).
Each disk has a diameter of $0.02$\,m and is positioned with a distance of $0.02$\,m between each adjacent pair.
We use two RealSense depth cameras (D$405$) to collect data.
Our nominal configuration set generated as described above contains $2773$ configurations.
For comparison purposes, we define a special $\mathcal{D}^{hys}$ subset, $\mathcal{D}^{hys}_{\mathrm{home}}$, as the first data collected where each $\vec{q}_\mathrm{prior}$ configuration is the home configuration.
The full data set $\mathcal{D}^{hys}$ consists of $\mathcal{D}^{hys}_{\mathrm{home}}$ augmented with two sets where $\vec{q}_\mathrm{prior}$ was instead randomized (as detailed above).
In this way, $\mathcal{D}^{hys}_{\mathrm{home}}$ contains $2773$ data points and $\mathcal{D}^{hys}$ contains $2773 \times 3 = 8319$ data points.
With this, each nominal configuration appears in our data set three times, once where the prior configuration was the home configuration and two additional times where the prior configuration was randomly selected.
Our method outputs a point cloud of size 512, and we downsample the point clouds of all training and evaluation data to have size 512 as well.

\subsection{Quantification of hysteresis}
We first quantify the severity of hysteresis for our robot.
To do so, we compare the nominal configuration set of $2773$ configurations gathered in two ways.
In the first we command the robot to return to the home configuration in between each of the nominal configurations (i.e., with the home configuration as $\vec{q}_\mathrm{prior}$).
In the second, we instead command the robot to visit a random configuration in between each of the nominal configurations (i.e., with a random configuration as $\vec{q}_\mathrm{prior}$).
We then compare the point clouds for each nominal configuration gathered in these two ways.
In the case where there is no hysteresis effect, these would result in the same shapes.
Instead, the difference between the two is a Chamfer distance of $0.04 \pm 0.1$\,m and an end-tip position distance of $0.018 \pm 0.013$\,m ($9 \pm 6.5\%$ of robot length), averaged over all 2773 configurations.
This implies that our tendon-driven robot displays significant hysteresis.
One example of the difference is shown in Fig.~\ref{fig:intro}, top right.

\subsection{Loss function evaluation}
\label{section: loss}

To evaluate the impact of our loss function $\mathcal{L}_{tendon}$, we train our method with a variety of loss functions and compare their performance.
Among various loss functions, in this work we compare our model trained with $\mathcal{L}_{tendon}$ against models trained with a standard Mean Squared Error (MSE) loss and Chamfer distance by itself as a loss function.
In Fig.~\ref{fig:loss_functions} we show examples of the models' outputs evaluated on one hysteresis configuration as a qualitative example.

Quantitatively, we evaluate the outputs of the model on our $50$ test cases.
The model trained with the MSE loss fails to model the robot's shape, achieving an average Chamfer distance of $1.134 \pm 0.167$\,m from the ground truth point clouds.
While the Chamfer only loss function produces point clouds that start to approximate the robot's geometry, the resulting points are unevenly distributed along the shape.
The Chamfer distance for the model trained on this loss function is $0.0084 \pm 0.0065$\,m.
The model trained on our proposed loss function, EMD + Chamfer ($\mathcal{L}_{tendon}$), shows the best performance both qualitatively and quantitatively.
It achieves an average Chamfer distance of $0.0057 \pm 0.0024$\,m, displaying a precise estimation of the robot's shape with evenly distributed point clouds.
We find that our proposed loss function ($\mathcal{L}_{tendon}$) provides point clouds that are $199$ and $1.47$ times closer to the ground truth on average than the MSE and the Chamfer loss functions, respectively.
Particularly, the EMD loss plays a crucial role in promoting an even distribution within the point cloud, improving the overall quality of shape prediction. 
Without the EMD in the loss function, the model tends to concentrate the predicted points near the proximal end of the robot, where the robot moves less than near the tip, which is undesirable as we would like the model to predict a uniformly dense point cloud along the robot's length.

\subsection{Comparison to physics-based and non-hysteresis models}
\label{section: comparison}

We compare against both a physics-based model and a version of our model that does not account for hysteresis.
We do so in two experiments, the first where we evaluate on a large number of independent hysteresis configurations and the second in which we evaluate on a smaller set of cohesive, sequential trajectories in the robot's workspace.

\subsubsection{Comparison method details}

\begin{figure}[t]
    \centering
    \includegraphics[width=\linewidth]{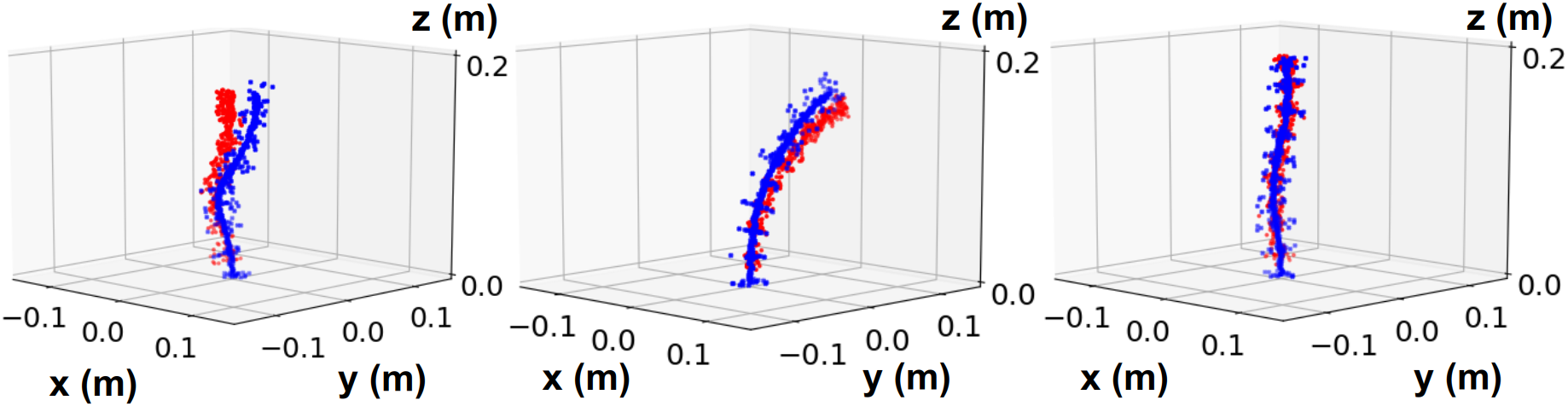}
  \caption{Physics-based predicted robot shape (blue) vs. ground truth (red). Three examples are shown from the test set, the worst (left), the median (middle), and the best (right). These demonstrate the ranges of error between the predicted shape and the ground truth sensed shape. 
  }
  \label{fig:sim_real_gap}
\end{figure}

We first compare against the state-of-the-art physics-based method of~\cite{Oliver-Butler2019ContinuumDisplacements}, which is an extension of the model proposed in~\cite{Rucker2011_TRO} to use tendon displacements rather than tension as the input.
We obtained an initial guess of the backbone Young's modulus E by hanging weights on one straight routed tendon, and optimizing E to best fit the shape of the manipulator. Then, we calibrated the model on a subset of 30 of the neural network training cases. The eight parameters we calibrated are the backbone Young's modulus $E$, Poisson's ratio $\nu$, mass density $\rho$, tendon compliance $C_t$ and tendon offsets $[\delta_1 , \ldots, \delta_4]$ which represent potential tension/slack in the home configuration. 
The physics-based model produces a backbone shape representation.
To generate a point cloud representation of the full robot from the model we radially expand the backbone shape to the diameter of the robot, attach the $9$ circular disks along the backbone in simulation, and create a pointcloud of the resulting geometry (see Fig.~\ref{fig:sim_real_gap}). We optimize the eight parameters using \textit{pattern search} to best match the collected pointclouds from the training set.

We also compare against a version of our model that does not account for hysteresis (which we label here as the non-hysteresis model).
To do so, we change the model's input from the two configuration pair (the hysteresis configuration) of our model to a single configuration---as in a traditional forward kinematics sense---while keeping the rest of the network architecture the same.
We train the model on the nominal configurations collected by returning to the home configuration in between each.

\subsubsection{Timing evaluation}
To demonstrate the value of our deep decoder network outputting the full shape of the robot directly, we first evaluate the time required by the various methods to compute the shape of the robot.
All computation was performed on a computer with an AMD Ryzen 7 3700X 8-core processor, 64 GB of 3200 Hz DDR4 DRAM, and an NVIDIA GeForce RTX 3060 Ti.

Our model requires a computation time of $0.32 \pm 0.00004$\,ms from the configuration input to outputting the full predicted robot shape, averaged across $100$ runs.
For the same evaluation, the non-hysteresis model demonstrates an average computation time of $0.33 \pm 0.00011$\, ms.
By contrast, it takes on average $613 \pm 206$\,ms to generate a predicted pointcloud from the physics-based model.
This computation time is broken up as $394$\,ms to get the backbone shape from the model and then an additional $219$\,ms to augment the backbone with the rest of the robot geometry's point cloud.
Our model demonstrates substantial speedup, producing full robot shapes $\approx 1,915$ times more quickly.

\subsubsection{Single hysteresis configuration evaluation}
\begin{figure}[t]
    \centering
    \includegraphics[width=0.90\linewidth]{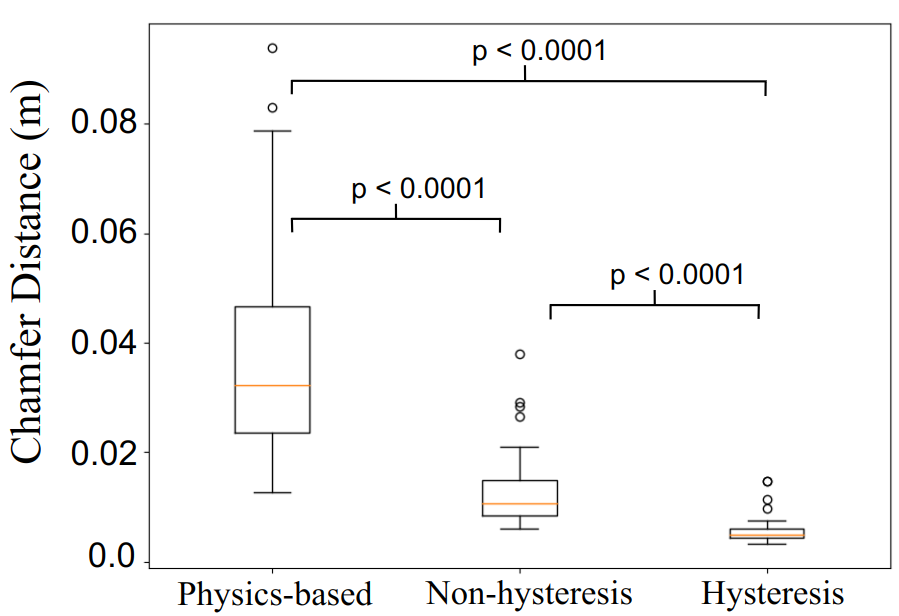}
  \caption{Boxplot comparison of the physics-based model, the non-hysteresis model, and our full model that accounts for hysteresis (labeled Hysteresis here) across the $50$ test cases. The physics-based model achieves an average Chamfer distance of $0.0374 \pm 0.019$\,m. The non-hysteresis model achieves a Chamfer distance of $0.013 \pm 0.007$\,m while our full model achieves $0.0057 \pm 0.0024$\,m.} 
  \label{fig:boxplots}
\end{figure}

\begin{figure}[t]
    \captionsetup{skip=-0.2pt}
   \centering
   \begin{subfigure}[b]{\linewidth}
        \centering
	\includegraphics[width=\textwidth]{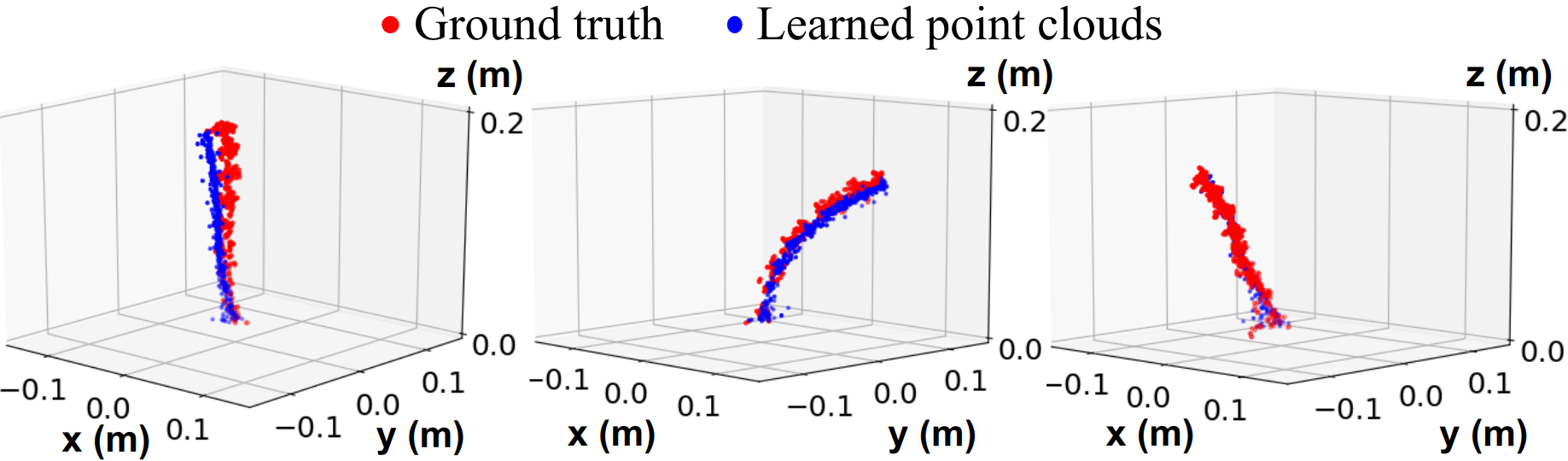}
    \caption{Non-hysteresis model predicted shape compared with the ground truth.}
    \label{fig:non_hys_on_hys_dataset}
  \end{subfigure}
  \begin{subfigure}[b]{\linewidth}
        \centering    
	\includegraphics[width=\textwidth]{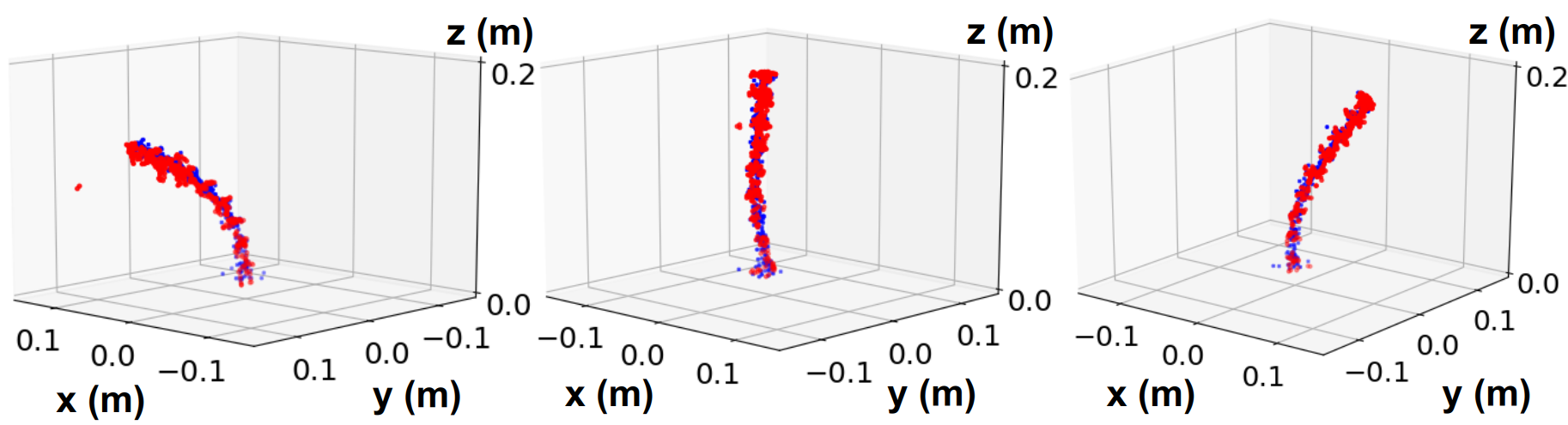}
    \caption{Full hysteresis model predicted shape compared with the ground truth.}
    \label{fig:hysteresis}
  \end{subfigure}
  \caption{Three examples (the worst, the median, and the best cases from the $50$ test cases) comparing the performance of the non-hysteresis (a) and our full model (b). 
  While the non-hysteresis model shows good qualitative performance, our full model estimates the robot's shape more accurately.
  }
  \label{fig:hysteresis_results}
\end{figure}

\begin{figure*}[t!]
    \captionsetup{skip=-0.1pt}
    \centering
    \includegraphics[width=0.9\linewidth]{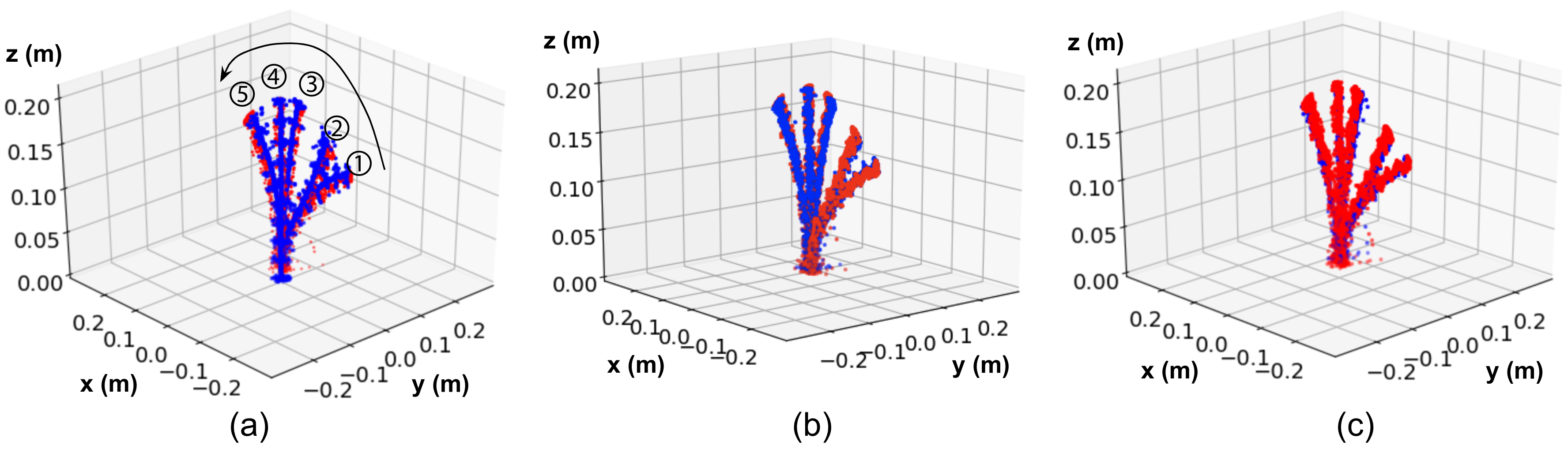}
  \caption{An example trajectory test case (the $1^{st}$ of the $6$ trajectory cases) comparing the performance of (a) the physics-based model, (b) the non-hysteresis model, and (c) our hysteresis model. The order of the robot trajectory is indicated in (a) using numbers 1 through 5, and this sequence is consistent across all three subfigures. The physics-based model performs the worst, with the non-hysteresis model improving upon the physics-based model, while our model shows improvement over both.}
  \label{fig:traj_plot}
  \vspace{-1.5em}
\end{figure*}

We next evaluate model accuracy.
For these experiments we employ the same $70\%$--$30\%$ training/validation set split during training but evaluate it on the separate $50$ test points from the full hysteresis data set used for all testing.

We show qualitative results for the physics-based model in Fig.~\ref{fig:sim_real_gap}.
We show quantitative results in Fig.~\ref{fig:boxplots} comparing all three methods with additional qualitative results for our model and the non-hysteresis model shown in Fig.~\ref{fig:hysteresis_results}.

Qualitatively (see Figs.~\ref{fig:sim_real_gap} and~\ref{fig:hysteresis_results}), the physics-based model performs the worst, the non-hysteresis model the next best, and our full model performing the best, more closely capturing  the robot's shape.
Quantitatively, as can be seen in Fig.~\ref{fig:boxplots}, when evaluated on our $50$ test data points our model with hysteresis significantly outperforms the other two.
The physics-based model performs the worst, resulting in an average Chamfer distance of $0.0374 \pm 0.019$\,m from the ground truth.
The non-hysteresis model achieves a Chamfer distance of $0.013 \pm 0.007$\,m while our hysteresis model achieves $0.0057 \pm 0.0024$\,m.
On average, our proposed method produces shapes that are $6.6$ times closer to the ground truth than the physics-based model and $2.3$ times closer than the non-hysteresis models.

As Chamfer distance, being a sum over many point pair distances, is a potentially unintuitive metric, we evaluate end-tip position error as an additional metric.
The physics-based model exhibits $0.019 \pm 0.008$\,m end-tip position error ($9.5 \pm 4\%$ of robot length), the non-hysteresis model $0.01 \pm 0.004$\,m ($5 \pm 2$\% of robot length), and our model $0.0076 \pm 0.0037$\,m ($3.8 \pm 1.85$\% of robot length).
However, note that Chamfer distance serves as a more precise metric than the end-tip position error, given that we calibrate/train all the models based on the complete shape of the robot, rather than solely focusing on the end-tip position error.

\subsubsection{Full trajectory evaluation}

For tendon-driven robots to be effectively used in real-world applications, it is imperative that the model accurately predicts the robot's shape as it executes trajectories in its environment.

In this section we evaluate our method's ability to predict the shape of the robot across full trajectories, composed of a number of sequentially chosen random configurations.

\begin{figure}[t!]
    \captionsetup{skip=-0.1pt}
    \centering
    \includegraphics[width=0.8\linewidth]{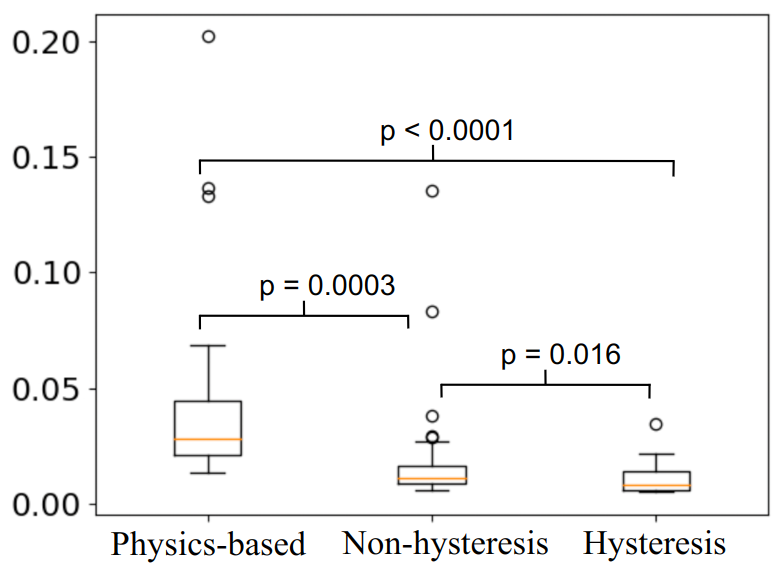}
  \caption{Boxplot comparison of Chamfer distance averaged across all configurations in all $6$ trajectories for the physics-based model, the non-hysteresis model, and our model (labeled Hysteresis here). The physics-based model achieves an average Chamfer distance of $0.044 \pm 0.042$\,m. The non-hysteresis model achieves $0.015 \pm 0.006$\,m while our model achieves $0.009 \pm 0.003$\,m.}
  \label{fig:traj_box}
\end{figure}

For this experiment we augment the model's training data by supplementing
$\mathcal{D}^{hys}$ with a random trajectory dataset consisting of $1200$ data points.
This dataset consists of configurations randomly selected from the nominal set without returning to $\vec{q}_{home}$ between configurations or between pairs of configurations.

We evaluate our model's performance compared with the physics-based model and the non-hysteresis model using $6$ trajectories not seen during training, each 
comprising $5$ different sequential randomly-sampled configurations.
Qualitatively, as can be seen in Fig.~\ref{fig:traj_plot}, our model accurately estimates the robot's shape for the entire trajectory, outperforming the other comparison methods.
Fig.~\ref{fig:traj_box} shows the quantitative result for all three methods.
The physics-based model achieves a Chamfer distance of $0.044 \pm 0.042$\,m from the ground truth averaged over all $6$ trajectory cases (i.e., $30$ configurations).
The non-hysteresis model displays an average Chamfer distance of $0.015 \pm 0.006$\,m, performing better than the physics-based model.
Our model exhibits the best performance with an average Chamfer distance of $0.009 \pm 0.003$\,m, producing shapes that are $4.9$ and $1.7$ times closer to the ground truth on average than the physics-based and non-hysteresis models, respectively.
The end-tip position error measurement shows $0.02 \pm 0.012$\,m ($10 \pm 6$\% of robot length) for the physics-based model, $0.011 \pm 0.004$\,m ($5.5 \pm 2$\% of robot length) for the non-hysteresis model, and $0.009 \pm 0.004$\,m ($4.5 \pm 2$\% of robot length) for our model.

\section{Conclusion}

In this work, we propose a deep decoder neural network model that computes the forward kinematics of tendon-driven continuum robots.
The model, trained with a novel loss function, directly outputs a full shape representation of the robot in the form of a point cloud.
Further, by augmenting the input with a prior configuration in addition to the current configuration, the model is able to successfully account for hysteresis.
We evaluate the accuracy of our method in predicting the robot's shape by using Chamfer distance as a metric and compare against a physics-based model and a model that does not account for hysteresis, significantly outperforming both.

In future work, we plan to develop a new approach for forward kinematics learning for tendon-driven robots that accounts for the entire past trajectory of the robot rather than a short history.
We also intend to leverage the model in control and planning for these robots, in methods such as, e.g.,~\cite{Bentley2022_IA,Huang2021_ISMR}.
We also intend to apply the method to other continuum robots, such as concentric tube robots.
We also plan to consider the potentially rate-dependent nature of hysteresis.

We also note that point clouds as an output geometric shape representation are popular in other domains but are not the only choice for geometric shape representation.
For instance, voxel sets and signed distance functions, among others, are options as well.
We intend to further explore the ways in which a learned model can output the robot's shape directly using representations such as these.

We believe that this new approach takes steps toward overcoming the limitations of current approaches in modeling the shape of tendon-driven robots.
This has the potential to lead to more accurate and efficient prediction of the tendon robot's shape, unlocking the door to safer operation and improved capabilities for these robots.

\bibliographystyle{IEEEtran}
\bibliography{2023-iros-cho}

\begin{thebibliography}{10}
\providecommand{\url}[1]{#1}
\csname url@rmstyle\endcsname
\providecommand{\newblock}{\relax}
\providecommand{\bibinfo}[2]{#2}
\providecommand\BIBentrySTDinterwordspacing{\spaceskip=0pt\relax}
\providecommand\BIBentryALTinterwordstretchfactor{4}
\providecommand\BIBentryALTinterwordspacing{\spaceskip=\fontdimen2\font plus
\BIBentryALTinterwordstretchfactor\fontdimen3\font minus
  \fontdimen4\font\relax}
\providecommand\BIBforeignlanguage[2]{{%
\expandafter\ifx\csname l@#1\endcsname\relax
\typeout{** WARNING: IEEEtran.bst: No hyphenation pattern has been}%
\typeout{** loaded for the language `#1'. Using the pattern for}%
\typeout{** the default language instead.}%
\else
\language=\csname l@#1\endcsname
\fi
#2}}

\bibitem{Dupont2022_ProcIEEE}
P.~Dupont, N.~Simaan, H.~Choset, and C.~Rucker, ``Continuum robots for medical
  interventions,'' \emph{Proceedings of the IEEE}, vol. 110, no.~7, pp.
  847--870, July 2022.

\bibitem{Burgner2015_TRO}
J.~Burgner-Kahrs, D.~C. Rucker, and H.~Choset, ``Continuum robots for medical
  applications: a survey,'' \emph{IEEE Transactions on Robotics}, vol.~31,
  no.~6, pp. 1261--1280, Dec. 2015.

\bibitem{Nguyen2015_IROS}
T.-D. Nguyen and J.~Burgner-Kahrs, ``A tendon-driven continuum robot with
  extensible sections,'' in \emph{{IEEE}/{RSJ} {International} {Conference} on
  {Intelligent} {Robots} and {Systems} ({IROS})}, Sept. 2015, pp. 2130--2135.

\bibitem{Kato2015_TMECH}
T.~Kato, I.~Okumura, S.-E. Song, A.~J. Golby, and N.~Hata, ``Tendon-driven
  continuum robot for endoscopic surgery: preclinical development and
  validation of a tension propagation model,'' \emph{IEEE/ASME Transactions on
  Mechatronics}, vol.~20, no.~5, pp. 2252--2263, Oct. 2015.

\bibitem{Rucker2011_TRO}
D.~C. Rucker and R.~J. Webster, III, ``Statics and dynamics of continuum robots
  with general tendon routing and external loading,'' \emph{IEEE Transactions
  on Robotics}, vol.~27, no.~6, pp. 1033--1044, Dec. 2011.

\bibitem{Oliver-Butler2019ContinuumDisplacements}
K.~Oliver-Butler, J.~Till, and C.~Rucker, ``{Continuum Robot Stiffness under
  External Loads and Prescribed Tendon Displacements},'' \emph{IEEE
  Transactions on Robotics}, vol.~35, no.~2, 2019.

\bibitem{Grassmann2018_IROS}
R.~Grassmann, V.~Modes, and J.~Burgner-Kahrs, ``Learning the forward and
  inverse kinematics of a 6-{DOF} concentric tube continuum robot in {SE}(3),''
  in \emph{{IEEE}/{RSJ} {International} {Conference} on {Intelligent} {Robots}
  and {Systems} ({IROS})}, 2018, pp. 5125--5132.

\bibitem{Kuntz2020_TMRB}
A.~Kuntz, A.~Sethi, R.~J. Webster, III, and R.~Alterovitz, ``Learning the
  {Complete} {Shape} of {Concentric} {Tube} {Robots},'' \emph{IEEE Transactions
  on Medical Robotics and Bionics}, vol.~2, no.~2, pp. 140--147, 2020.

\bibitem{Yu2018_MS}
H.-J. Yu, W.-L. Yang, Z.-X. Yang, W.~Dong, Z.-J. Du, and Z.-Y. Yan,
  ``Hysteresis analysis of a notched continuum manipulator driven by tendon,''
  \emph{Mechanical Sciences}, vol.~9, no.~1, pp. 211--219, June 2018.

\bibitem{Kim2021_arXiv}
Y.-H. Kim and T.~Mansi, ``Shape-adaptive hysteresis compensation for
  tendon-driven continuum manipulators,'' \emph{arXiv preprint
  arXiv:2109.06907}, 2021.

\bibitem{Wu2019_CVPR}
W.~Wu, Z.~Qi, and L.~Fuxin, ``{PointConv}: {Deep} {Convolutional} {Networks} on
  {3D} {Point} {Clouds},'' in \emph{Proceedings of the {IEEE}/{CVF}
  {Conference} on {Computer} {Vision} and {Pattern} {Recognition}}, 2019, pp.
  9621--9630.

\bibitem{Qi2017_CVPR}
C.~R. Qi, H.~Su, K.~Mo, and L.~J. Guibas, ``{PointNet}: {Deep} {Learning} on
  {Point} {Sets} for {3D} {Classification} and {Segmentation},'' in
  \emph{Proceedings of the {IEEE}/{CVF} {Conference} on {Computer} {Vision} and
  {Pattern} {Recognition}}, 2017, pp. 652--660.

\bibitem{Thach2022_ICRA}
B.~Thach, B.~Y. Cho, A.~Kuntz, and T.~Hermans, ``Learning {Visual} {Shape}
  {Control} of {Novel} {3D} {Deformable} {Objects} from {Partial}-{View}
  {Point} {Clouds},'' in \emph{{International} {Conference} on {Robotics} and
  {Automation} ({ICRA})}, May 2022, pp. 8274--8281.

\bibitem{Bentley2022_IA}
M.~Bentley, C.~Rucker, and A.~Kuntz, ``Interactive-{Rate} {Supervisory}
  {Control} for {Arbitrarily}-{Routed} {Multitendon} {Robots} via {Motion}
  {Planning},'' \emph{IEEE Access}, vol.~10, pp. 80\,999--81\,019, 2022.

\bibitem{Swaney2016_JMD}
P.~J. Swaney, P.~A. York, H.~B. Gilbert, J.~Burgner-Kahrs, and R.~J. Webster,
  ``Design, fabrication, and testing of a needle-sized wrist for surgical
  instruments,'' \emph{Journal of Medical Devices}, vol.~11, no.~1, p. 014501,
  Mar. 2017.

\bibitem{Leavitt2023_Hamlyn}
A.~Leavitt, R.~Lam, N.~C. Taylor, D.~S. Drew, and A.~Kuntz, ``Toward a
  millimeter-scale tendon-driven continuum wrist with integrated gripper for
  microsurgical applications,'' \emph{arXiv preprint arXiv:2302.07252}, 2023.

\bibitem{Yuan2019_MMT}
H.~Yuan, L.~Zhou, and W.~Xu, ``A comprehensive static model of cable-driven
  multi-section continuum robots considering friction effect,'' \emph{Mechanism
  and Machine Theory}, vol. 135, pp. 130--149, May 2019.

\bibitem{Poignonec2020_RAL}
T.~Poignonec, P.~Zanne, B.~Rosa, and F.~Nageotte, ``Towards {In} {Situ}
  {Backlash} {Estimation} of {Continuum} {Robots} {Using} an {Endoscopic}
  {Camera},'' \emph{IEEE Robotics and Automation Letters}, vol.~5, no.~3, pp.
  4788--4795, July 2020.

\bibitem{Kato2016_IJCARS}
T.~Kato, I.~Okumura, H.~Kose, K.~Takagi, and N.~Hata, ``Tendon-driven continuum
  robot for neuroendoscopy: validation of extended kinematic mapping for
  hysteresis operation,'' \emph{International Journal of Computer Assisted
  Radiology and Surgery}, vol.~11, no.~4, pp. 589--602, Apr. 2016.

\bibitem{baek2020hysteresis}
D.~Baek, J.-H. Seo, J.~Kim, and D.-S. Kwon, ``Hysteresis compensator with
  learning-based hybrid joint angle estimation for flexible surgery robots,''
  \emph{IEEE Robotics and Automation Letters}, vol.~5, no.~4, pp. 6837--6844,
  2020.

\bibitem{guo2023motion}
Y.~Guo, B.~Pan, Y.~Sun, G.~Niu, Y.~Fu, and M.~Q.-H. Meng, ``Motion hysteresis
  compensation based on motor current segmentation for elongated cable-driven
  surgical instruments,'' \emph{IEEE Transactions on Automation Science and
  Engineering}, 2023.

\bibitem{Neppalli2009_AR}
S.~Neppalli, M.~A. Csencsits, B.~A. Jones, and I.~D. Walker, ``Closed-form
  inverse kinematics for continuum manipulators,'' \emph{Advanced Robotics},
  vol.~23, no.~15, pp. 2077--2091, 2009.

\bibitem{wu2021hysteresis}
D.~Wu, Y.~Zhang, M.~Ourak, K.~Niu, J.~Dankelman, and E.~Vander~Poorten,
  ``Hysteresis modeling of robotic catheters based on long short-term memory
  network for improved environment reconstruction,'' \emph{IEEE Robotics and
  Automation Letters}, vol.~6, no.~2, pp. 2106--2113, 2021.

\bibitem{bai2021task}
W.~Bai, F.~Cursi, X.~Guo, B.~Huang, B.~Lo, G.-Z. Yang, and E.~M. Yeatman,
  ``Task-based lstm kinematic modeling for a tendon-driven flexible surgical
  robot,'' \emph{IEEE Transactions on Medical Robotics and Bionics}, vol.~4,
  no.~2, pp. 339--342, 2021.

\bibitem{Jung2011_IROS}
J.~Jung, R.~S. Penning, N.~J. Ferrier, and M.~R. Zinn, ``A modeling approach
  for continuum robotic manipulators: {Effects} of nonlinear internal device
  friction,'' in \emph{{IEEE}/{RSJ} {International} {Conference} on
  {Intelligent} {Robots} and {Systems}}, Sept. 2011, pp. 5139--5146.

\bibitem{Subramani2015_ICRA}
G.~Subramani and M.~R. Zinn, ``Tackling friction - an analytical modeling
  approach to understanding friction in single tendon driven continuum
  manipulators,'' in \emph{{IEEE} {International} {Conference} on {Robotics}
  and {Automation} ({ICRA})}, May 2015, pp. 610--617.

\bibitem{Lilge2022_IJRR}
S.~Lilge, T.~D. Barfoot, and J.~Burgner-Kahrs, ``Continuum robot state
  estimation using gaussian process regression on se (3),'' \emph{The
  International Journal of Robotics Research}, vol.~41, no. 13-14, pp.
  1099--1120, 2022.

\bibitem{Xu2017_IJMRCAS}
W.~Xu, J.~Chen, H.~Y. Lau, and H.~Ren, ``Data-driven methods towards learning
  the highly nonlinear inverse kinematics of tendon-driven surgical
  manipulators,'' \emph{The International Journal of Medical Robotics and
  Computer Assisted Surgery}, vol.~13, no.~3, p. e1774, 2017.

\bibitem{Liang2021_ICRA}
N.~Liang, R.~M. Grassmann, S.~Lilge, and J.~Burgner-Kahrs, ``Learning-based
  {Inverse} {Kinematics} from {Shape} as {Input} for {Concentric} {Tube}
  {Continuum} {Robots},'' in \emph{{IEEE} {International} {Conference} on
  {Robotics} and {Automation} ({ICRA})}, May 2021, pp. 1387--1393.

\bibitem{Bergeles2015_Hamlyn}
C.~Bergeles, F.~Y. Lin, and G.~Z. Yang, ``Concentric tube robot kinematics
  using neural networks,'' in \emph{Hamlyn symposium on medical robotics}, June
  2015, pp. 1--2.

\bibitem{Fagogenis2016_IROS}
G.~Fagogenis, C.~Bergeles, and P.~E. Dupont, ``Adaptive nonparametric kinematic
  modeling of concentric tube robots,'' in \emph{{IEEE}/{RSJ} international
  conference on intelligent robots and systems ({IROS})}, Oct. 2016, pp.
  4324--4329.

\bibitem{Nair2020_ICML}
V.~Nair and G.~E. Hinton, ``Rectified linear units improve restricted boltzmann
  machines,'' in \emph{Proceedings of the 27th international conference on
  machine learning (ICML-10)}, 2010, pp. 807--814.

\bibitem{Ioffe2015_ICML}
S.~Ioffe and C.~Szegedy, ``Batch {Normalization}: {Accelerating} {Deep}
  {Network} {Training} by {Reducing} {Internal} {Covariate} {Shift},'' in
  \emph{Proceedings of the 32nd {International} {Conference} on {Machine}
  {Learning}}, June 2015, pp. 448--456.

\bibitem{Arun1987_TPAMI}
K.~S. Arun, T.~S. Huang, and S.~D. Blostein, ``Least-squares fitting of two
  3-{D} point sets,'' \emph{IEEE Transactions on pattern analysis and machine
  intelligence}, no.~5, pp. 698--700, 1987.

\bibitem{Glorot2010_AISTATS}
X.~Glorot and Y.~Bengio, ``Understanding the difficulty of training deep
  feedforward neural networks,'' in \emph{Proceedings of the {Thirteenth}
  {International} {Conference} on {Artificial} {Intelligence} and
  {Statistics}}.\hskip 1em plus 0.5em minus 0.4em\relax JMLR Workshop and
  Conference Proceedings, Mar. 2010, pp. 249--256.

\bibitem{Huang2021_ISMR}
Y.~Huang, M.~Bentley, T.~Hermans, and A.~Kuntz, ``Toward learning
  context-dependent tasks from demonstration for tendon-driven surgical
  robots,'' in \emph{International Symposium on Medical Robotics (ISMR)}, Nov.
  2021.

\end{thebibliography}

\end{document}